\documentclass[conference]{IEEEtran}
\IEEEoverridecommandlockouts
\usepackage{cite}
\usepackage{amsmath,amssymb,amsfonts}
\usepackage{algorithmic}
\usepackage{graphicx}
\usepackage{textcomp}
\usepackage{xcolor}

\usepackage{booktabs}
\usepackage{multirow}
\usepackage{pifont}
\usepackage{threeparttable}

\usepackage{soul, color, xcolor}
\usepackage{float}

\makeatletter
\newcommand{\biggg}{\bBigg@{1.2}}  
\def\bigggl{\mathopen\biggg}

\makeatother

\def\BibTeX{{\rm B\kern-.05em{\sc i\kern-.025em b}\kern-.08em
		T\kern-.1667em\lower.7ex\hbox{E}\kern-.125emX}}
\begin{document}
	\title{{\vspace{-3ex} \normalsize \textsuperscript{*}2024 IEEE 40th International Conference on Data Engineering (ICDE)} \vspace{2.5ex} \\Accurate Explanation Model for Image Classifiers \\using Class Association Embedding\\
		\thanks{\textsuperscript{*}Camera-ready version of ICDE 2024 accepted paper.}
		\thanks{\textsuperscript{$\dag$}Corresponding authors.}
	}
	
	\author{\IEEEauthorblockA{Ruitao Xie\textsuperscript{1, 2}, Jingbang Chen\textsuperscript{1}, Limai Jiang\textsuperscript{1, 2}, Rui Xiao\textsuperscript{1}, Yi Pan\textsuperscript{1, 3$\dag$}, Yunpeng Cai\textsuperscript{1$\dag$}}
		\IEEEauthorblockA{\textit{\textsuperscript{1}Shenzhen Institute of Advanced Technology, Chinese Academy of Sciences, Shenzhen 518055, China} \\
			\textit{\textsuperscript{2}University of Chinese Academy of Sciences, Beijing 100049, China}\\
			\textit{\textsuperscript{3}Shenzhen Key Laboratory of Intelligent Bioinformatics, Shenzhen 518055, China
			}
			\\
			\{rt.xie, jb.chen, lm.jiang2, rui.xiao, yi.pan, yp.cai\}@siat.ac.cn}}
	
	
\maketitle

\begin{abstract}
	Image classification is a primary task in data analysis where explainable models are crucially demanded in various applications. Although amounts of methods have been proposed to obtain explainable knowledge from the black-box classifiers, these approaches lack the efficiency of extracting global knowledge regarding the classification task, thus is vulnerable to local traps and often leads to poor accuracy. In this study, we propose a generative explanation model that combines the advantages of global and local knowledge for explaining image classifiers. We develop a representation learning method called class association embedding (CAE), which encodes each sample into a pair of separated class-associated and individual codes. Recombining the individual code of a given sample with altered class-associated code leads to a synthetic real-looking sample with preserved individual characters but modified class-associated features and possibly flipped class assignments. A building-block coherency feature extraction algorithm is proposed that efficiently separates class-associated features from individual ones. The extracted feature space forms a low-dimensional manifold that visualizes the classification decision patterns. Explanation on each individual sample can be then achieved in a counter-factual generation manner which continuously modifies the sample in one direction, by shifting its class-associated code along a guided path, until its classification outcome is changed. We compare our method with state-of-the-art ones on explaining image classification tasks in the form of saliency maps, demonstrating that our method achieves higher accuracies. The class-associated manifold not only helps with skipping local traps and achieving accurate explanation, but also provides insights to the data distribution patterns that potentially aids knowledge discovery. The code is available at https://github.com/xrt11/XAI-CODE.
\end{abstract}

\begin{IEEEkeywords}
	image classifier, explainable artificial intelligence, class association embedding, counterfactual generation, class-associated codes
\end{IEEEkeywords}

\section{Introduction}
Image classifiers are widely used in various data engineering tasks such as image retrieval, searching, captioning or annotation, data augmentation, as well as essential steps in question answering [1]-[4]. Image classification is also a primary task in data analysis with extensive applications such as face recognition, medical diagnosis, object recognition, fault detection, etc.  In the past decade, the developing of deep neural networks significantly boost the performance of image classification models. Nevertheless, the “black-box” nature of deep neural networks [5], namely, the lack of explainable inferences, creates a major and sustained challenge. It has been recognized that unexplainable models suffer from the risks of systematic defects, such as short-cut learning [6], and vulnerability to adversarial attacks [7], which are unbearable for many critical applications, such as medical diagnosis [8]. Hence, in recent years, there has been increasing interest in deriving explanations from machine learning models or developing intrinsic interpretable models [9], both of which fall into the category of explainable artificial intelligence (XAI). Notably, a fairly large portion of XAI approaches originates from image classifiers.

Typically, XAI methods can be classified into global and local methods, where global methods seek an explanation applicable across the entire data domain, while local methods strive to find detailed attributions for individual samples. Most popular XAI methods fall into local approaches, where key features associated with the classification outcomes are identified and a weight is assigned to each feature. Typical local methods include gradient-based [10]-[14] attributions, perturbation-based [15]-[18] attributions, and counterfactual generation methods [19]-[21]. These methods are straightforward to understand and typically accurate in explaining individual instances. However, due to the pervasive nonlinearity and complex interactions between features and context, local attributions are highly vulnerable to local traps, resulting in biased [22], unstable [23], inaccurate, or even misleading [24], [25] explanations. Moreover, it is persistently criticized [26], [27] that feature associations alone provide poor semantic knowledge about the decision rules. Global explainable methods, on the other hand, provide a complete picture of how decision patterns exist in the data domain. Nevertheless, the picture is mostly achieved by a simplified approximation of the deep learning models, resulting in severe performance degradations. For this reason, global methods are usually imprecise in explaining the detailed decision attribution for a given instance, which is often demanded in applications such as lesion or abnormality localization, making them less popular. To overcome the shortcomings of current methods and achieve accurate XAI, a model that explains a system of quantitative rules learned from the entire dataset but can vividly apply to each sample would be favorited [28], [29]. However, methods that can exploit the advantages of both global and local knowledge are lacking.

In this study, to address this issue, we develop a representation learning method called class association embedding (CAE). The CAE learns a separate paired class-individual subspace representation in which the class-associated subspace provides a concise encoding of sample features relevant to the classification task, while the individual subspace encodes sample-specific features that are conditionally independent of the classes. In contrast to previous global methods which irreversibly simplifies the data space, the embedding created by CAE can be reversely decoded to accurately reconstruct the original data. Furthermore, by manipulating the class-associated code we can perform direct modification on one sample to change its class-associated features and flip its classification outcomes, but preserve its individual characters and remain real-looking. We propose a building-block coherency feature extraction (BBCFE) method for efficiently gathering class-associated features in a low-dimensional space. Unlike existing feature extraction approaches that attempt to maximize the classification accuracy in the extracted feature space, BBCFE strives to minimize class-associated features in the individual code subspace so that the class-associated code subspace contains a comprehensive but concise mapping of the classification rules.

By adopting CAE and BBCFE, we well extract the global knowledge across the entire dataset, represented by a class-discriminative and rule-visualizing manifold, and, on this basis, develop a new methods for accurately explaining the behavior of black-box image classifiers on the data. Specifically, for each sample instance we want to explain, we first encode it in the class-associated space, then create a guided transition path starting from its class-associated code to the counter class(es) within this manifold. By combining the individual code of the sample with the class-associated codes sampled along this path, a series of synthetic samples with preserved background characteristics but changed class-associated features are generated. Local explanations (in the form of saliency maps) can be obtained by comparing these synthetic samples, which attributes the original features that mostly affect the classifier. With the assistance of the global knowledge and guided paths provided by the class-associated manifold, the generated local explanations efficiently skip the misleading local traps and thus achieve higher accuracy. 

Our contributions can be summarized as below:
\begin{itemize}
	\item We propose a class association embedding framework with building-block coherency feature extraction training method, which successfully separates class-associated features from class-unassociated features, and describes the decision behavior of a black-box classifier in a low-dimensional manifold. The manifold can help with discovering global classification knowledge and data patterns within the dataset. 
	\item We develop an instance-based explaining model for image classifiers by improving counterfactual generation with the guide of the knowledge and path plans provided by the class-associated manifold, which produces more accurate saliency map on localizing attributed features for individual images.
	\item We perform experiments on five datasets to validate the effectiveness and generalization ability of our algorithm. As a result, more effective and rule-compliant global knowledge is learned and extracted from the datasets, and with the help of the global knowledge, our framework achieves better performance compared with current state-of-the-art XAI methods on explaining classifiers derived from these datasets. We also conducts a series of explorations to demonstrate various properties of the global manifold we have learned, which includes semantic pervasiveness, semantic coherancy, smoothness and aiding in global understanding of classifier behavior.
\end{itemize}

\section{Related work}
\subsection{Local explanations}
\textbf{Gradient based methods}
Gradient-based methods posits that the gradient of the output with respect to each input unit can serve as an indicator of the significance of the input unit. Class Activation Mapping (CAM) [30] is the most famous gradient-based explainable algorithm, where only the last convolutional layer of the network could be analyzed and global average pooling layer must be included in the network, which is not conducive to generalization. GradCAM and GradCAM++ [11], [12] have been proposed as enhancements, enabling the analysis of any convolutional layers and without imposing strict requirements on the network architecture. This increased flexibility allows for broader applicability. Fullgrad [31] aggregates information from the input-gradient and all intermediate bias-gradients, which demarcates the objects satisfactorily, while Simple Fullgrad and Smooth Fullgrad [32] are its two variants. Token semantic coupled attention map (TS-CAM) [33] introduces visual transformer [34] and splits the image into multiple patch tokens to make different tokens aware of object categories for generating accurate attention maps. Gradient-based XAI methods are widely recognized and promoted due to their simplicity. Nevertheless, gradient-based techniques possess certain limitations, such as the issue of gradient vanishing [35] and their vulnerability to misleading local traps.
  
\textbf{Perturbation based methods}
The perturbation-based approaches aim to modify the input, such as selectively masking pixels in an image, in order to alter the output results. By observing the significant changes in output results caused by different modifications, the importance of various elements can be measured. LIME [15], a well-known perturbation method, generates new data by perturbing the input and trains a simple and interpretable model to approximate the output of a complex model. Several local perturbation methods have been derived from LIME, including [18], [36]. Researchers have proposed many other perturbation methods [37]-[40]. However, these methods lack guidance and meaningful perturbations, require extensive computational resources, and suffer from the loss of real distribution in samples generated by replacing the observed parts using heuristic (like ‘0’) filling methods. Additionally, local perturbations do not take contextual information into account.

\textbf{Counterfactual generation based methods}
Recent studies [19]-[21], [41]-[44] have suggested an approach called counterfactual generation [19], [20], [45] to explain machine learning models by modifying semantic-level features (such as super pixels or word terms) rather than original variables using generative models, such as generative adversarial networks (GANs), and observing how the prediction results shift with modifications. Few studies [46], [47] have proposed finding a combination of semantic features (called “fault lines”) that best push the samples toward the model decision boundary. However, these methods mostly rely on predefined labels and semantic terms [43], [48], [49] (like pseudo lesion segmentation maps required in LAGAN [49]) and generate modified samples with local random walks [20], [50], [51], which are unable to capture global knowledge, thus resulting in very limited improvements in explanation accuracy compared with previous methods. ICAM-reg [51] is currently the only method that also propose a attribute latent subspace, however, we’ll show in our experiment that it does not learn the classification rules in the subspace.

\subsection{Trap problems for local methods}
\begin{figure}
	\centering
	\includegraphics[scale=0.33]{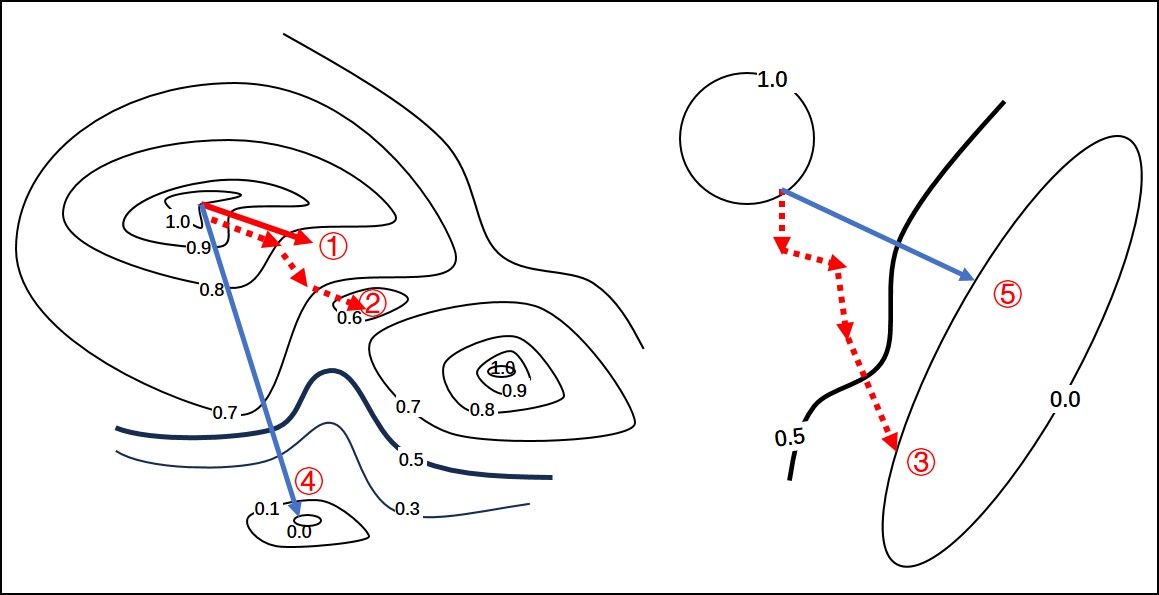}
	\caption{Illustration of traps for local explaining methods on a two-dimensional plane. The digits on the contour lines shows the classification probabilities and the black bold contour line shows the class-flipping border. \ding{172} Gradient-based or single perturbation methods may find misleading explanation directions. \ding{173} multi-perturbation methods (including counter-factual generation) tend to get trapped into local optima. \ding{174} Fault-line counter generations, without knowing the global picture, generate far-from-closest paths that lead to false-positive attributions. \ding{175} \ding{176} blue arrows show the correct explanation paths.}
	\label{localtraps}
\end{figure}

The above-described local explanation methods adopt gradient descending or greedy random walks to identify the combination of factors that lead to steepest degradation of classification probabilities, so that the most relevant features can be found. However, due to the complexity of real-world problems. The decision function contour can be highly non-linear and multi-peaked, so that the local gradient/descending direction may be divergent from the real class-flipping direction (where the desired attribution factor exists). Hence gradient-based, perturbation-based, or multi-perturbation counterfactual generation methods would be often trapped in local optima. Although fault-line based counterfactual generation methods search for the class-flipping points in the function plain to avoid local traps, without knowing the global picture, they generate a lot of unnecessary moves and resulted in an end point that is far from optimal, which also result in false-positive modifications and thus wrong attributions. Both cases are depicted in Fig. \ref{localtraps}.

\subsection{Global explanations}
Most global explainable methods aim to achieve explainability by constructing simple and transparent models, such as rule sets, trees, or equations [52]-[57]. These models possess simplistic architectures and a reduced number of parameters, leading to subpar predictive capabilities when confronted with intricate nonlinear problems. Some researchers have proposed knowledge distillation techniques [58]-[64], wherein the predictions of complex models are utilized as soft targets to train simple and explainable models. These approaches brings improved performance of simple models. However, due to their reliance on intricate models as instructors, these simple models encounter difficulties in attaining the same level of proficiency as their instructors. 

This paper presents an explainable approach utilizing counterfactual generation, which not only attains more precise local explanations but also accomplishes global explanations without compromising the performance of the original intricate model.

\section{METHODOLOGY}
\subsection{Problem Definition}
Explainable learning works on a black-box model $b(x)=y$ with a series of input-output pairs $\{x_n,y_n\}_{(n=1,..., N)}$ and $y \in \mathbb{C}$ by generating an explanation ${\Psi}(x,y)$ for the black-box model. The form of $\Psi$ is versatile; however, in most existing methods, $\Psi (x_n,y_n)$ includes a saliency map, indicating the importance of variables in $x_n$, which leads to the classification decision $y_n$. $\Psi$ can be either model-dependent or model-agnostic. Our method falls into the latter category in the sense that it requires only $\{x_n,y_n\}_{(n=1,..., N)}$ as inputs without knowing the details of $b$.

\subsection{Class Association Embedding Using a Cyclic Generative Adversarial Network}

It is straightforward that a low-dimensional representation of ${\Psi}(x,y)$, if applicable, would be favored because humans are weak in understanding high-dimensional topologies. However, previous attempts to explain black-box models in low-dimensional space usually lead to poor performance owing to the intrinsic complexity of both the classification task and the deep neural networks. In this study, we argue that this complexity can be compressed by separating class-common and individual-specific representations because a large portion of the variations exhibited in the tasks are induced by the morphing of semantically identical objects (i.e., specific patterns) under diverse backgrounds (e.g., different sizes, locations, orientations, and colors). By adopting GAN technology [65], we can decompose the data manifold into a class-associated and individual-specific subspace that can be freely combined to reconstruct various valid (real-looking) samples following the same distribution as the real dataset. The decomposition allows for a simplified encoding of the class-associated subspace because the individual backgrounds are removed.

\begin{figure*}
	\centering
	\includegraphics[scale=0.36]{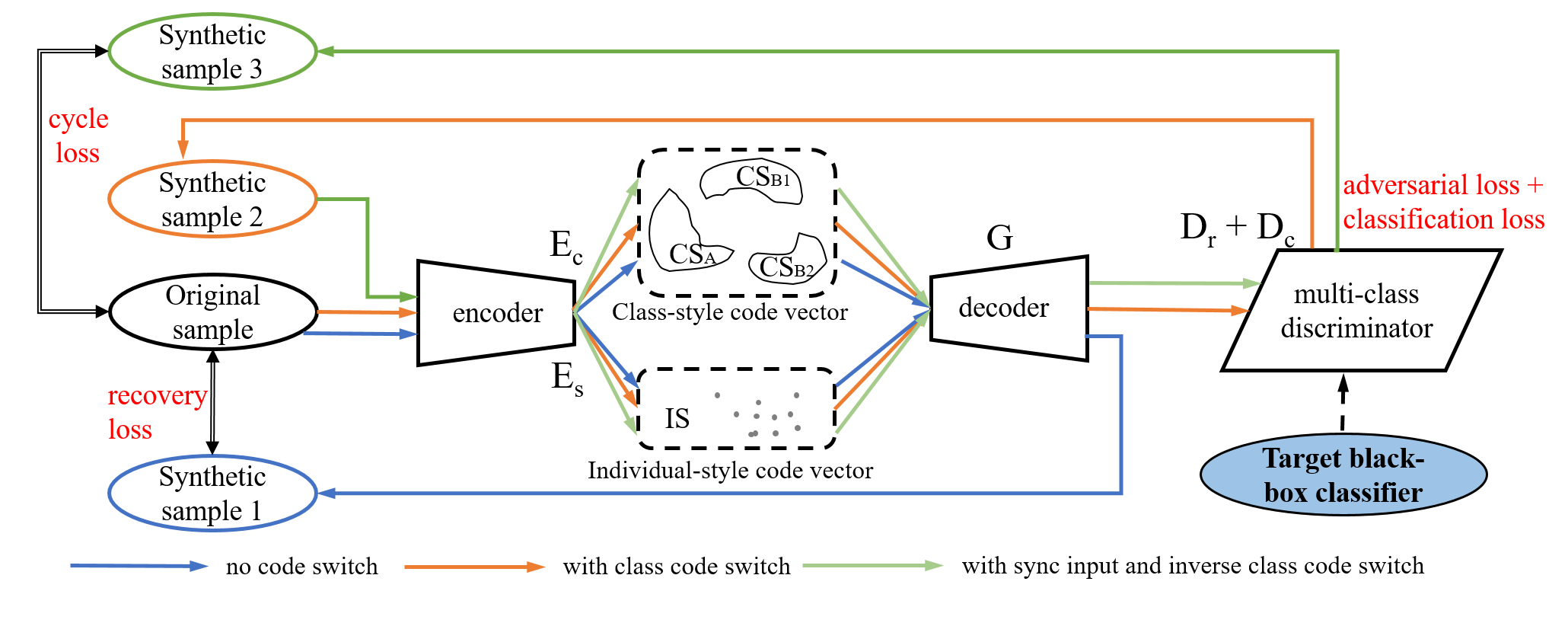}
	\caption{The overall framework of class association embedding. The recovered sample (generated from the original sample) and the recovered synthetic sample (generated from the synthetic sample) are expected to resemble the original sample, while the synthetic sample (generated from the original sample) is expected to inherit the individual style of the original sample (class $A$) but with the classification features of class $B1$ or $B2$ (which deceives the target classifier).}
	\label{overall-framework}
\end{figure*}

The proposed class association embedding framework is shown in Fig. \ref{overall-framework}, which is in the form of a cyclic generation adversarial network (cycle-GAN), including an encoder $E$, a decoder $G$, and a multiclass discriminator $D$. The encoder comprises two modules: $E_c$, which encodes class-associated style ($CS$, $c$ for simplicity) codes, and $E_s$, which encodes individual style ($IS$, $s$ for simplicity) codes. $CS$ codes are dedicated to extracting class-associated information from images (such as moustaches and lipstick in face images for gender classification task) and eliminating class-irrelevant information (such as the outline of the face, background, and glasses in face images for gender classification task). On the contrary, IS codes are dedicated to extracting class-irrelevant information from images and eliminating class-associated information. The decoder uses both the class-associated and individual code vectors as inputs to generate a new synthetic sample. The discriminator concurrently generates two outputs that attempt to discriminate between the real and synthetic samples ($Dr$) while determining the proper class assignments ($Dc$). The target black-box classifier can be also directly used as $Dc$. Unlike most existing cycle-GANs [66]-[68], which are asymmetric in two directions, we adopt a symmetric structure to encode all sample classes in the same unified space to explain the relationships between different classes.

\subsection{Learning the Separation of Code Spaces with Building-Block Coherency Feature Extraction}
The most essential step in CAE is to efficiently learn the low-dimensional representation of class-associated features and separate them from individual features. Previous image-to-image translation approaches have adopted the so-called disentangling technique, which selects class-relevant latent features from a generated candidate set by flipping them individually and observing the outcomes [50], [69]. However, disentangling is inefficient because of the following defects: relying on how the latent features are generated, ignoring the interaction between features, and often leading to a large feature set. 

\begin{figure*}
	\centering
	\includegraphics[scale=0.438]{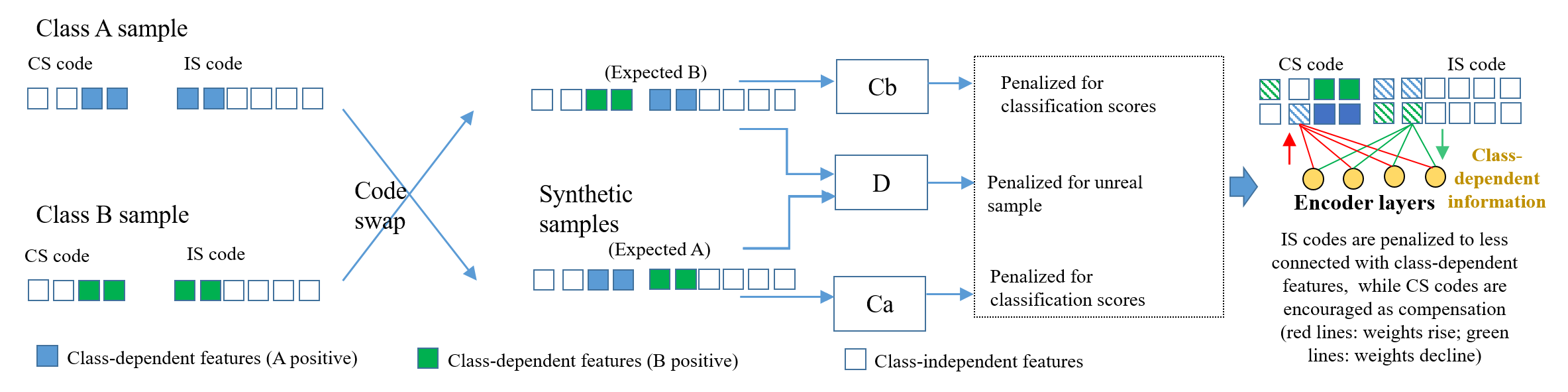}
	\caption{Illustration of learning for the separation of code spaces using building-block coherency feature extraction method.}
	\label{training-PBBFE}
\end{figure*}

To learn a low-dimensional representation of the class-associated features more efficiently, we propose the building-block coherency feature extraction (BBCFE) supervised training method as illustrated in Fig. \ref{training-PBBFE}. The basic idea of BBCFE is that there are many statistical dependencies between class-associated features. If some class-associated features are encoded in the CS code but some remain in the IS code, and we swap the CS code for a pair of encoded samples with different classes, we will result in a pair of chimeric samples that possess the features of both classes. There would be a higher chance that the chimeric samples look unreal and confuse the classifier than nature samples. By doing a large number of pair swapping, we can then identify the class-associated features encoded in the IS code space and weaken it through penalization, while on the other hand encouraging the emergence of class-associated features in the CS code space.

Suppose that we have m samples of class $A$ and n samples of class $B$, BBCFE enumerates all m $\times$ n pairs by feeding each pair of samples into the same encoder to obtain their class-associated and individual codes, respectively. A code-swapping scheme is performed so that the recombination codes can be decoded into two synthetic samples with switched class assignments to each other. A second-round combination is performed by switching the class-associated code back, and the new synthetic samples are expected to recover their original class assignments. If the dataset is large, we randomly sample from the m $\times$ n combinations to approximate the data distribution while reducing computational costs. A given sample $x_A$ in class $A$ is combined with numerous samples from class $B$. Let $(c_A, s_A)$ be the decomposed code of $x_A$. Swapping $c_A$ with the code of another sample $x_B$ from class $B$ yields $(c_B, s_A)$. Owing to the classification penalty, if $(c_B, s_A)$ does not decode to a synthesis sample of class $B$, it is penalized. Hence, $c_B$ is initially trained to contain class-associated features that serve as seeds for the building blocks. If $s_A$ contains class-associated information that is rarely present in class $B$, $(c_B, s_A)$ (probabilistically) leads to an unreal sample, which is penalized by the discriminator $D_r$. By penalizing this type of pair-incoherency between $c_B$ and $s_A$ over a large set of sample pairs, the class-associated features contained in the individual style space vanish. Then, to maintain the reconstruction accuracy, the vanished features are induced into the class-associated space as compensation using the shared latent layers in the encoded network. In this way, class-associated features accumulate and fuse in the class-associated code space as the training proceeds. Although BBCFE does not penalize a useless feature in the class-associated space, its low-dimensional property naturally imposes $\ell_0$-analog regularization to remove useless features. Unlike disentangling, BBCFE prefers to select a strong building-block pattern set rather than a set of scattered variables; thus, it is more efficient for extracting global knowledge.

\begin{figure}
	\centering
	\includegraphics[scale=0.23]{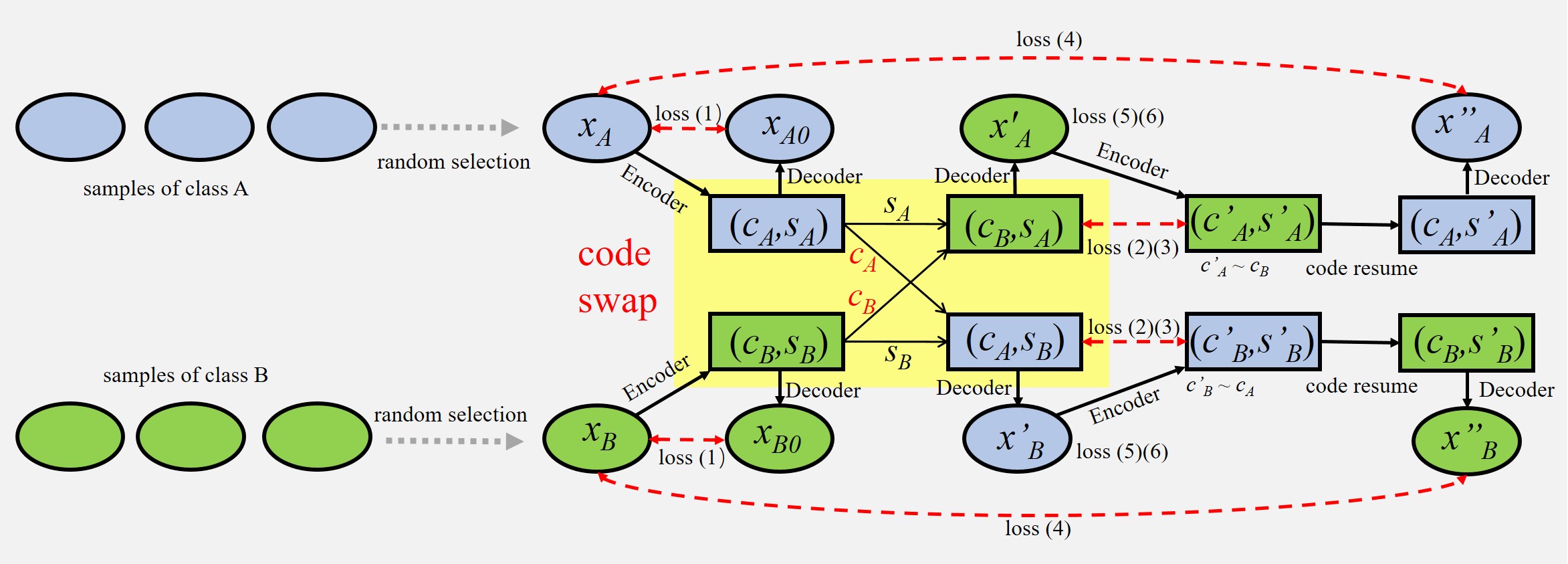}
	\caption{The detailed training schema for Building-Block Coherency Feature Extraction.}
	\label{training}
\end{figure}

The detailed training schema for Building-Block Coherency Feature Extraction is presented in Fig. \ref{training}. The numbers in parentheses showed the portion of loss functions given by the corresponding equation in Section 3.D. As is pointed out, existing cyclic adversarial learning schemes do not fit in our architecture. In order to learn the efficient representation of the class-associated style in a unified domain, we randomly select two images from different classes from the training set for pairing. Without loss of generality, for two paired samples $x_A$ and $x_B$, the two classes can be denoted by $y_A$ and $y_B$, respectively. Multi-class tasks can be also learned in a 1-vs-1 manner. The paired samples ($x_A$ and $x_B$) are fed into the same encoder to generate encodings ($c_A$, $s_A$) and ($c_B$, $s_B$) respectively, where $c$ represents the class-associated codes while $s$ denotes the individual codes. 
To successfully embed class-associated information of the paired sample and achieve class transferring, a code-swap scheme (swapping CS codes) is performed so that the combination $(c_B, s_A)$ and $(c_A, s_B)$ are fed into the decoder for generating samples $x_A'$ and $x_B'$ with expected switched class assignments $y_A'=y_B$ and $y_B'=y_A$. Both samples are then re-encoded as $(c_A',s_A')$ and $(c_B',s_B')$ with $c_A' \sim c_B$ and $c_B' \sim c_A$. A second-round combination is peformed and thus $(c_A, s_A')$ and $(c_B, s_B')$ are decoded into $x_A'' \sim x_A$ and $x_B'' \sim x_B$, respectively. During random paired training process, without loss of generality, a sample can be randomly (or enumeratedly when the data size is small) paired with many samples of different classes to generate a large number of class-individual code combinations. After enough iterations of training, class-associated and individual subspaces are separated successfully.

\subsection{Design of Loss Functions}
Loss functions are designed to efficiently learn the class association embedding vectors and achieve cyclic adversarial learning goals. Let $A : (x_A, y_A)$ be a sample to be credited and $B : (x_B, y_B)$ be its paired sample (whose loss is credited separately). Let $E_{c}$, $E_{s}$, and $G$ be the mapping functions of the encoders and decoders, as described above. The first part of the loss function is the reconstruction loss. The basic reconstruction loss, as shown in equation \ref{con:loss1}, represents the error of the encode–decode loop without any CS-code swap. Moreover, we argue that modifying the class-associated code of a sample does not affect its individual code, and vice versa. Hence, we encode the synthetic sample and compare its class-associated and individual codes with their original counterparts, resulting in the class-wise and sample-wise reconstruction losses denoted in equations \ref{con:loss2} and \ref{con:loss3}, which ensures the semantic uniqueness and unity of the embedding manifold, and helps to improve the consistency of the code space during training. It is noteworthy that the adoption of equations \ref{con:loss2} and \ref{con:loss3} is crucial for achieving the homeomorphic (topology-maintaining) property of the embedded space. For most cycle-GAN approaches, recovery losses are imposed on only the image, or on only the image and the class-associated code, which leads to unsmooth code mapping, thereby distorting the topological structures. 

\begin{equation}
	\mathcal{L}_{recon}^{x_{A}}=\mathbb{E}_{x_{A} \sim p(x_{A})}\left[ {\parallel G(E_{c}(x_{A}),E_{s}(x_{A}))-x_{A} \parallel}_{1} \right]
	\label{con:loss1}
\end{equation}
\begin{equation}
	\mathcal{L}_{recon}^{c_{A}}=\mathbb{E}_{c_{A} \sim p(c_{A}), s_{B} \sim q(s_{B})} \left[ {\parallel E_{c}(G(c_{A},s_{B}))-c_{A} \parallel}_{1} \right]
	\label{con:loss2}
\end{equation}
\begin{equation}
	\mathcal{L}_{recon}^{s_{A}}=\mathbb{E}_{c_{B} \sim p(c_{B}), s_{A} \sim q(s_{A})} \left[ {\parallel E_{s}(G(c_{B},s_{A}))-s_{A} \parallel}_{1} \right]
	\label{con:loss3}
\end{equation}

The second part of the loss function is the cyclic loss, as shown in equation \ref{con:loss4}, which calculates the error of recovering the original sample after the two-round encode-decode cycle with two CS-code swappings. The third and fourth parts are the adversarial and classification loss functions, as shown in equations \ref{con:loss5} and \ref{con:loss6}, respectively, where $D_{r}$ denotes the output probability of the sample being real looking ($1$ refers to real, whereas $0$ refers to fake), and $D_{c}$ denotes the probability of assigning the correct class ($y_B$).

\begin{equation}
	\mathcal{L}_{cyc}^{x_{A}}=\mathbb{E}_{x_{A} \sim p(x_{A})}\left[ {\parallel G(c_{A}, E_{s}(G(c_{B}, s_{A}) ))-x_{A} \parallel}_{1} \right]
	\label{con:loss4}
\end{equation}
\begin{equation}
	\mathcal{L}_{adv1}^{A2B}=-\mathbb{E}_{c_{B} \sim p(c_{B}), s_{A} \sim q(s_{A})} log(\frac{e^{D_{r}(G(c_{B},s_{A}))[1]}}{\sum_{i=0}^{i=1}e^{D_{r}(G(c_{B},s_{A}))[i]}})
	\label{con:loss5}
\end{equation}

\begin{equation}
	\mathcal{L}_{cla1}^{A2B}=-\mathbb{E}_{c_{B} \sim p(c_{B}), s_{A} \sim q(s_{A})} log(\frac{e^{(D_{c}(G(c_{B},s_{A}))[y_{B}])}}{\sum_{j \in \mathbb{C}} e^{(D_{c}(G(c_{B},s_{A}))[j])}})
	\label{con:loss6}
\end{equation}

Combining the above loss functions (with $\lambda_{1}$, $\lambda_{2}$, $\lambda_{3}$, $\lambda_{4}$, $\lambda_{5}$ and $\lambda_{6}$ as weights), we used the object function in equation \ref{con:loss7} to train and update the encoder and decoder.

\begin{equation}
	\begin{split}
		\mathcal{L}(E, G)=\lambda_{1}(\mathcal{L}_{recon}^{x_{A}}+\mathcal{L}_{recon}^{x_{B}})+\lambda_{2}(\mathcal{L}_{recon}^{c_{A}}+\mathcal{L}_{recon}^{c_{B}})\\
		+\lambda_{3}(\mathcal{L}_{recon}^{s_{A}}+\mathcal{L}_{recon}^{s_{B}})+\lambda_{4}(\mathcal{L}_{cyc}^{x_{A}}+\mathcal{L}_{cyc}^{x_{B}})\\+\lambda_{5}(\mathcal{L}_{adv1}^{A2B}+\mathcal{L}_{adv1}^{B2A})
		+\lambda_{6}(\mathcal{L}_{cla1}^{A2B}+\mathcal{L}_{cla1}^{B2A})
		\label{con:loss7}
	\end{split}
\end{equation}

However, the parameters in discriminator $D$ are updated based on equations \ref{con:loss8}, \ref{con:loss9} and \ref{con:loss10}, which are calculated for each pair of samples together. During training, both real and generative images are fed into the adversarial discriminator, but only real images are fed for classification. Hence, parameters $\varphi_{1}$ and $\varphi_{2}$ are used to balance the two losses.

\begin{equation}
	\begin{split}
		\mathcal{L}_{adv2}^{A2B}=-\mathbb{E}_{c_{B} \sim p(c_{B}), s_{A} \sim q(s_{A})} log(\frac{e^{D_{r}(G(c_{B},s_{A}))[0]}}{\sum_{i=0}^{i=1}e^{D_{r}(G(c_{B},s_{A}))[i]}})\\-\mathbb{E}_{x_{B} \sim p(x_{B})} log(\frac{e^{(D_{r}(x_{B})[1])}}{\sum_{i=0}^{i=1}e^{D_{r}(x_{B})[i]}})
		\label{con:loss8}
	\end{split}
\end{equation}
\begin{equation}
	\mathcal{L}_{cla2}^{A}=-\mathbb{E}_{x_{A} \sim p(x_{A})} log(\frac{e^{D_{c}(x_{A})[y_{A}]}}{\sum_{j \in \mathbb{C}} e^{D_{c}(x_{A})[j]}})
	\label{con:loss9}
\end{equation}
\begin{equation}
	\mathcal{L}(D)=\varphi_{1}(\mathcal{L}_{adv2}^{A2B}+\mathcal{L}_{adv2}^{B2A})+\varphi_{2}(\mathcal{L}_{cla2}^{A}+\mathcal{L}_{cla2}^{B})
	\label{con:loss10}
\end{equation}

\subsection{Local Explanations by Guided Counterfactual Generation on the Class-Associated Manifold}
\begin{figure}
	\centering
	\includegraphics[scale=0.238]{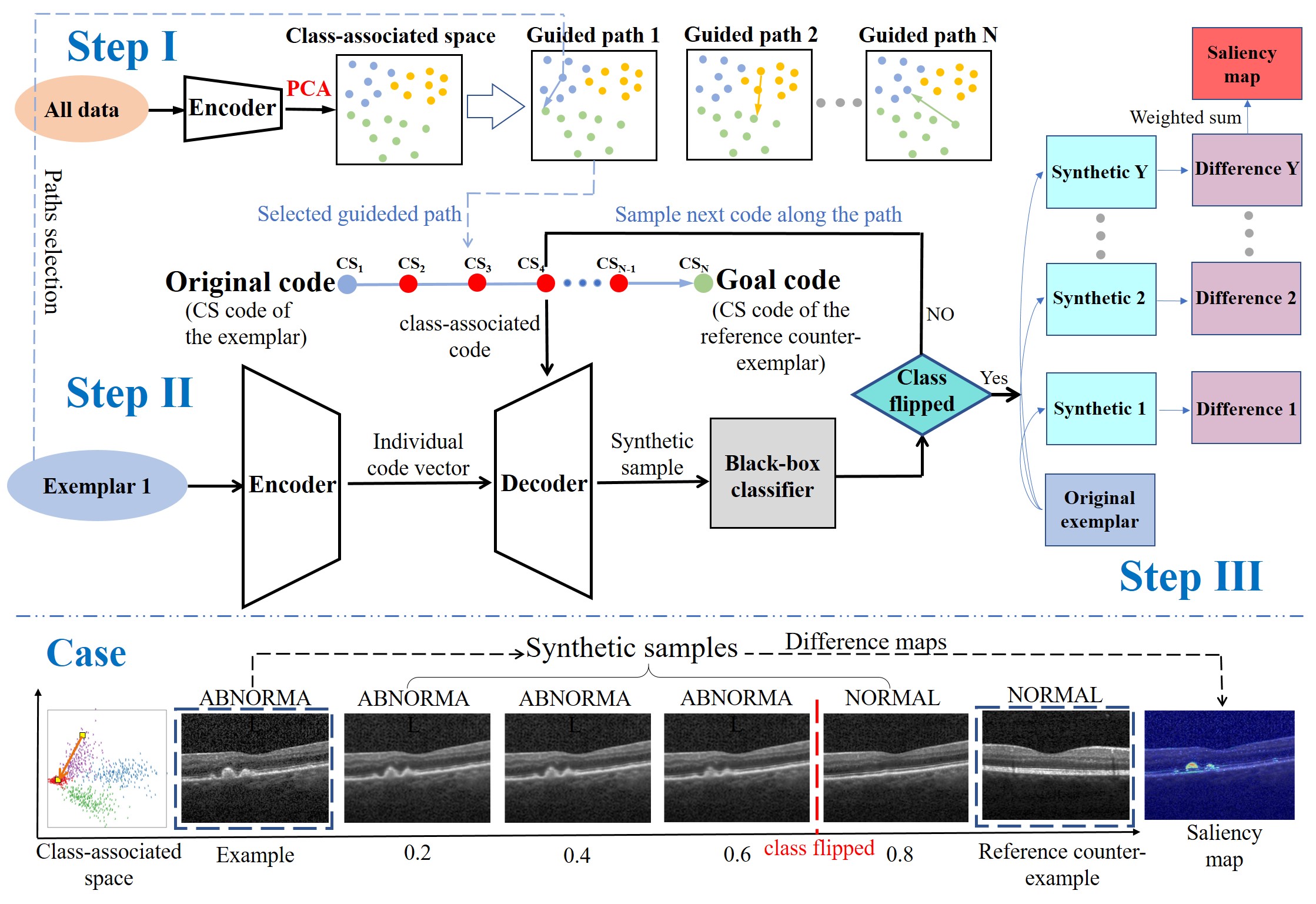}
	\caption{XAI framework based on class association embedding and generation model. In the class-associated space, codes from samples with different classes are mapped into dots with different colors.}
	\label{XAI-framework}
\end{figure}

Using the proposed class association embedding method, we learn a separate representation of class-associated and individual features for each sample. The manifold learned in the class-associated code space provides a representation of the global explanation of the data distribution and behavior of the target model. Analytic skills, such as topological or correlation analysis, with external variables, can be adopted on the embedded manifold to enable knowledge discovery. We apply the manifold on individual samples to generate accurate local explanations. The framework is depicted in Fig. \ref{XAI-framework}. First, the entire dataset is fed into the CAE network to learn the class-associated manifold, which can be visualized by adopting low-dimensional projection techniques, such as principal component analysis (PCA) or t-distributed stochastic neighbor embedding (t-SNE) [70]. Any given sample (including new samples not in the training set) to be explained (called an exemplar) can thus be mapped to a location in the space. Second, a class transition path is plotted by connecting the sample location to the counter class. A series of class-associated codes can be obtained via linear interpolation on the path, which is then combined with the individual code of the exemplar to generate synthetic samples. The generation process can be stopped if the predicted class of the currently generated sample by the classifier changes to the target class as expected. Or sometimes for convenience, we can directly use the class-associated code of the target/counter sample for generation. Finally, we compare the generated sample series to find a concise attribution to explain the behavior of the target classifier. The sampled images are aligned, and then the frame-to-frame differences are calculated, resulting in a series of differential maps. A saliency map can be obtained by summing all these differential maps with weights proportional to their classification probability changes or, in many cases, simply by contrasting the destination image to the original image if the selected transition path is linear. In the case of multiclass tasks, multiple paths, and thus multiple saliency maps, can be created to explain different class-to-class contrast features.

Following the above steps, we locate the regions of interest (ROI) for images for the entire class transition path, which is insusceptible to local traps or noises. Thus, this surpasses existing methods that adopt only local perturbation or pairwise contrasts, leading to more accurate and detailed saliency maps. Moreover, users can interactively design a transition path to serve various knowledge discovery purposes, such as mining different pathology rules or contrasting intraclass subtypes. We can use only one exemplar (or a few exemplars) to demonstrate the entire set of transition paths by traversing the class-associated space, providing a perceptual understanding of the global behavior of the target model, which is unachievable using current XAI methods.

\section{Experiments}
\subsection{Dataset and Experiments Settings}
We utilized a total of five datasets to validate the performance of our algorithm, including OCT [71], Brain Tumor1 [72], Brain Tumor2 [73], Chest X-rays [71], Human Face [74] datasets. Considering the significance of explainability in computer-aid diagnosis systems, we employed four medical imaging datasets, encompassing three commonly-used imaging methods: Optical Coherence Tomography imaging [71], Magnetic resonance imaging [72], [73] and X-ray imaging [71]. These datasets covered three prevalent disease types that exert a substantial burden on healthcare system and cause distress to patients: retinal diseases [71], pneumonia [71], and brain tumors [72], [73]. To evaluate the generalization ability of our method, we also conducted experiments on Human Face dataset [74] (human face images were commonly employed in generation research about attribute transformation). The numbers of images in the training and testing sets for five datasets were presented in TABLE \ref{data-number}. In the OCT dataset, abnormal images encompassed three types of lesions: Choroidal Neovascularization (CNV), Diabetic Macular Edema (DME), and DRUSEN. The training set comprised 8000 images for each of the three diseases, while the test set consisted of 250 images for each disease. The Chest X-rays datasets categorized abnormalities as pneumonia, while in the Brain Tumor1 and Brain Tumor2 datasets, abnormal samples referred to brain images with tumor.

%
%
%
%
%
%
%
%

\begin{table*}
	
	\caption{The numbers of used images in training and testing sets in five datasets.}
	\label{data-number}
	\centering
		\scalebox{1}{
		\begin{tabular}{lcccccccc}
			\toprule
			Datasets  &OCT     &Brain Tumor1  &Brain Tumor2  &Chest X-rays  &\multicolumn{2}{c}{Human Face} \\
			\midrule
			
			Train set (normal)   &8000  &1200 &710 &1349 &(female) &23243 \\
			
			\midrule
			Train set (abormal)  &24000  &1200 &4398 &3883 &(male) &23766 \\
			
			\midrule
			
			Test set (normal)   &250  &300 &302 &234 &(female) &5841 \\
			
			\midrule
			Test set (abormal)  &750  &300 &1623 &390 &(male) &5808 \\
			
			\midrule
			Classification task &retinal disease  &brain tumor &brain tumor &pneumonia &\multicolumn{2}{c}{gender} \\

			\bottomrule
		\end{tabular}
	}
\end{table*}

In our experiments, the input images were center cropped to a square and resized to 256 $\times$ 256 (or directly resized as 256 $\times$ 256). A random horizontal flip with a probability of 0.5 for input images was taken for data enhancement during training. We used an identical hyperparameter set for all tasks. The class-associated code was set to 8 d, and the individual code was 256 $\times$ 64 $\times$ 64 d. The Adam optimizer parameters were set to 1e-4 for both the initial learning rate and weight decay. The loss function weights were set as $\lambda_{1}=10$, $\lambda_{2}=1$, $\lambda_{3}=1$, $\lambda_{4}=10$, $\lambda_{5}=1$, $\lambda_{6}=1$, $\varphi_{1}=1$ and $\varphi_{2}=2$, thus, the accuracy of image reconstruction was emphasized to enable high-quality image-to-image contrasts. 

\subsection{Baseline Methods}
We compared our method with 9 state-of-the-art explaining methods:

(1)\textbf{Perturbation Methods}: We compared the LIME [15] algorithm, which is the most classic local-perturbation method for XAI, and often used as a baseline for comparison on XAI studies in recent years. 

(2)\textbf{Gradient Methods}: There are many existing gradient-based XAI algorithms, which were widely used due to their simplicity. We compared five gradient-based XAI algorithms as following. The classical Grad-CAM [11] utilized the activation values of convolutional layers for explaining black-box models, which was widely used as a baseline. Fullgrad [31] utilized activation values, gradient values, and bias values from all layers, achieving good performance for XAI, while Simple Fullgrad and Smooth Fullgrad [32] were its two variants. TS-CAM [33] generated class activation maps using vision transformer, achieving state-of-the-art performance.

(3)\textbf{Counterfactual generation Methods}: Counterfactual generation for XAI research has emerged in recent years. Masking key areas and changing key features based on latent space were two main strategies used for counterfactual generation. LAGAN [49] belongs to the former strategy, while StyLEx [50] and ICAM-reg [51] belongs to the latter. Both methods were recently published with SOTA performance.

\subsection{Evaluation Metrics}
We compared our method with several state-of-the-art XAI methods in terms of the accuracy of generating saliency maps for explanations. To quantitatively evaluate the accuracy of the generated saliency maps, we adopted the evaluation criteria proposed in [75], [76], which sorts the pixels based on their saliency values and covers the most important pixels (with a 7 $\times$ 7 patch centered on each pixel and filled by random values) to see how the classification performance degrades with the growth of the covered areas. 
For each sample image, we calculated the predicted probability changes of the classifier for the ground truth class with the growth of the covered areas. Let $K$ be the number of samples, $N$ the total number of coverages evaluated, and $p=1..N$ the number of currently covered patches. The average decreasing value of the class probability for all images with $p$ covered patches is called the overall degradation at $p$. Two evaluation metrices are then introduced. The area-over-the-perturbation-curve (AOPC), as expressed in equation \ref{con:loss11}, is the average of overall degradations for all $p=1..N$, where $X_{i, p}$ denotes the $i$-th image with $p$ top patches covered (depend on the xAI model), and $F_{c}(X_{i, p})$ represents the probability of $X_{i, p}$ for the ground truth class predicted by the black-box classifier. The peak degradation (PD), on the other hand, formulated in equation \ref{con:loss12}, is the maximum overall degradation for all $p$.

\begin{small}
\begin{equation}
	AOPC =\sum_{p=1}^{p=N}\sum_{i=1}^{i=K}[F_{c}(X_{i,0})-F_{c}(X_{i,p})] \bigggl / (K \times N)
	\label{con:loss11}
\end{equation}
\end{small}

\begin{small}
\begin{equation}
	PD =MAX \left(\sum_{i=1}^{i=K}[F_{c}(X_{i,0})-F_{c}(X_{i,p})] \bigggl / K\right)_{p=1,2..,N}
	\label{con:loss12}
\end{equation}
\end{small}

For each dataset we trained a ResNet50 [77] classifier as the external black-box model to explain. Although our method could be adopted on various models, many of the baseline methods described below relied on computations inside the classifier and were tuned on CNN-based models, hence using ResNet50 would be suitable for the baselines. All the methods used the same target black-box classifier (except for TS-CAM [33], which created its classifier rather than explaining external ones).

\subsection{Performance Comparison on Generating Saliency Map Explanations}
We randomly selected some test images from the OCT, Brain Tumor1, Brain Tumor2, Chest X-rays, and Human Face datasets for evaluating the performance of the generated saliency maps. 
The evaluation results for the attribution performance of generated saliency maps by all methods were presented in TABLE \ref{compared-evalutaion}. We observed that our methods show significantly higher peak degradation (PD) of the predicted probability of the ground-truth class and a higher area-over-the-perturbation-curve (AOPC) than all other methods in all five datasets, meaning that it provided more accurate and comprehensive explanations than previous XAIs. Furthermore, it is noteworthy that all baseline methods suffered from poor performance on at least two datasets, as measured by AOPC, suggesting that they were easily trapped by the locally deceptive patterns in the dataset and produced numerous false-positive features.

\begin{table*}
	
	\caption{QUANTITATIVE EVALUATION OF SALIENCY MAPS OBTAINED USING DIFFERENT XAIS ON BENCHMARK DATASETS.}
	\label{compared-evalutaion}
	\centering
		\scalebox{1}{
		\begin{tabular}{lcccccccccc}
			\toprule
			Methods    &LIME &Ful. &Sim.  &Smo.  &Gra. &StyLEx &TSC.  &LAG. &ICA. &CAE (ours)\\
			\midrule
			OCT (AOPC)   &0.513 &0.307 &0.302 &0.314 &0.365 &0.313 &0.337 &0.500 &0.490 & \textbf{0.571} \\
			OCT (PD)  &0.682 &0.454 &0.439 &0.461 &0.498 &0.551 &0.503 &0.669 &0.670 & \textbf{0.708}\\
			\midrule
			Brain Tumor1(AOPC)  &0.046 &0.092 &0.090 &0.079 &0.014 &0.071 &0.070 &0.105 &0.098 & \textbf{0.116} \\
			Brain Tumor1(PD) &0.067 &0.138 &0.136 &0.111 &0.032 &0.130 &0.092 &0.151 &0.153 & \textbf{0.182}\\
			\midrule
			Brain Tumor2(AOPC)  &0.387 &0.446 &0.466 &0.302 &0.267 &0.584 &0.504 &0.578 &0.510 & \textbf{0.605} \\
			Brain Tumor2(PD) &0.681 &0.582 &0.607 &0.448 &0.520 &0.673 &0.590 &0.663 &0.595 & \textbf{0.689}\\
			\midrule
			Chest X-rays(AOPC)  &0.695 &0.767 &0.782 &0.669 &0.485 &0.818 &0.767 &0.912 &0.883 & \textbf{0.917} \\
			Chest X-rays(PD) &0.931 &0.954 &0.961 &0.891 &0.769 &0.955 &0.928 &0.963 &0.961 & \textbf{0.968}\\
			\midrule
			Human Face(AOPC)  &0.261 &0.124 &0.127 &0.139 &0.101 &0.089 &0.145 &0.274 &0.222 & \textbf{0.293} \\
			Human Face(PD) &0.358 &0.238 &0.244 &0.263 &0.193 &0.212 &0.216 &0.405 &0.419 & \textbf{0.422}\\
			
			\bottomrule
		\end{tabular}
	}
\end{table*}

\begin{figure*}
	\centering
	\includegraphics[scale=0.245]{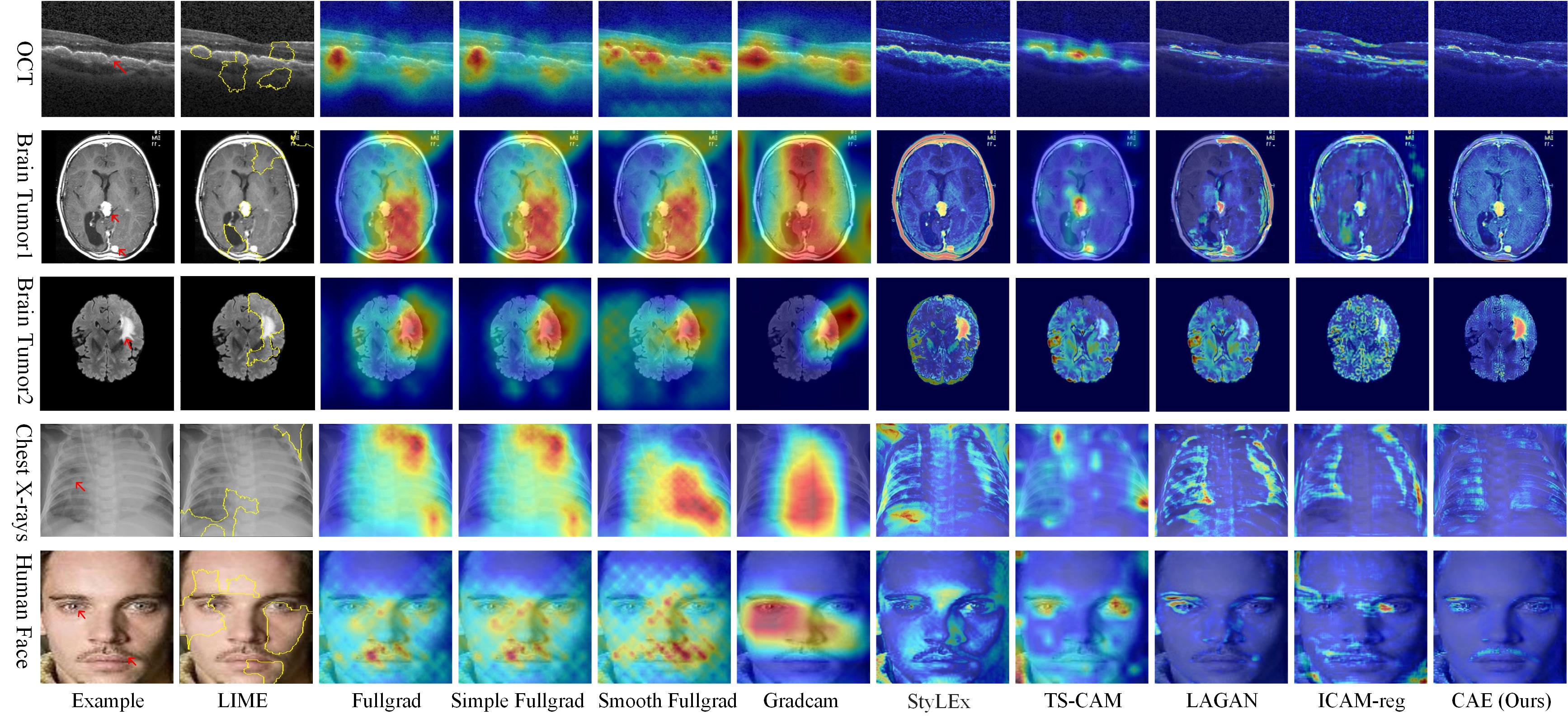}
	\caption{Comparison results of generated saliency maps using different methods on the OCT, Brain Tumor1, Brain Tumor2, Chest X-rays, and Human Face datasets. The ground truth is indicated by the red arrows in the first image.}
	\label{all-data-sal}
\end{figure*}

To demonstrate the powerful ability of our method in XAI more intuitively, we also showed examples of generated saliency maps using CAE and the baseline methods in Fig. \ref{all-data-sal}.  We can see that the regions of interest (ROIs, the parts highlighted) obtained were more accurate and finer grained with clearer contours when using our method. Compared with baseline methods, which were mostly based on local perturbations or gradients, our algorithm used global data distribution rules to achieve more sophisticated, accurate, and context-aware knowledge extraction with the aid of guided transition paths. Moreover, for medical images in which the classification patterns are highly diversified and subtle, existing methods often suffer from noises or background fluctuations and generate false ROIs (e.g., LIME), as shown in the figure. Our method is powerful in skipping these traps by removing the diverse background features and preferring collective building-block patterns.

\begin{figure}
	\centering
	\includegraphics[scale=0.185]{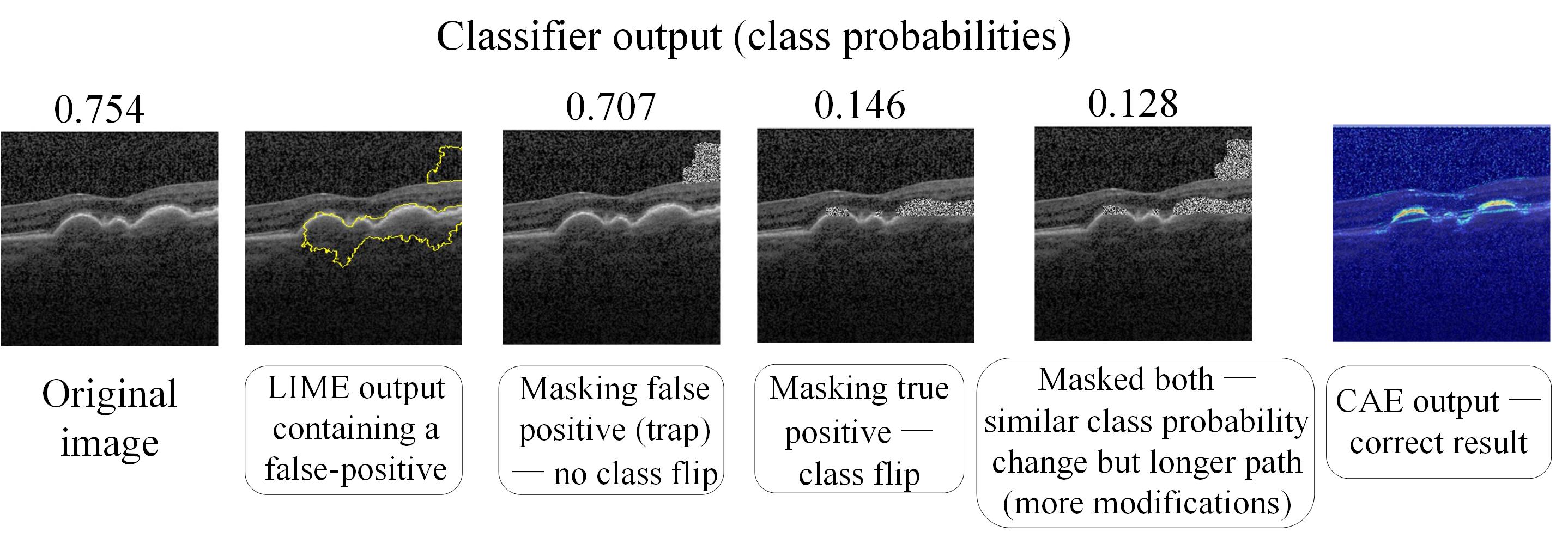}
	\caption{Example of a false positive generated by LIME due to a local trap. Masking the false positive does lead to local decreasing of the classification probability, which is able to deceive greedy-based methods. On the other hand, seeking global optimal path help with averting local traps (our method).}
	\label{local-traps}
\end{figure}

Fig. \ref{local-traps} gives an intutive example about the local trap problem and how our method averts it. The figure shows an OCT image with a lesion (true positive, mid-right), and a false positive identified by the baseline method LIME (top-right, which is completely out of the tissue region). We observe that, by masking the false positive region, the classification probability generated by the classifier does decrease, though not causing class flipping, which explains why gradient or greedy-based methods can be trapped. On the other hand, masking the true positive region leads to a sharp, class-flipping decrease of classificaiton probability. Furthermore, we see that masking both true and false positive will leads to similar probability drop as masking true positive only, but with a longer modification path (more regions covered). This explains that our method, by finding a nearly shortest class-flipping path with the aid of global class-associated knowledge, is more inclined to exclude false positives, thus skipping local traps.

\subsection{Comparison of Global Knowledge Extraction}
A prominent feature of our approach was to learn a low dimensional manifold that embeded the global trends of the classifier decision pattern. One may wonder that this could be also achieved using previous counter-factual generation methods, especially latent space approaches. Here we demonstrated that the global map obtained by our methods was not achievable by previous methods. We compared our class-associated code space with the “attribute latent space” of ICAM-reg, our closest peer methods, on how good they preserved the classification patterns. The most objective and undoubtful measurement was that how the test samples could be classified solely using their attribute latent code. To assess the significance of the difference between the two methods, we employed a ten-fold cross validation procedure so that not only the mean classification accuracies but also the standard deviations were obtained. Random forest classifier was adopted for both methods on all datasets with the same parameters. The resulted classification performances were presented in TABLE \ref{tsne-classification-analysis}. We see that, despite that ICAM-reg have strived to optimize the latent-space classification accuracy in training, our method drastically outperformed ICAM-reg in terms of classification accuracy and stability on all test datasets, proving that the class-associated manifolds extracted by our method have far better class-discriminative property.

\begin{table*}
	
	\caption{Comparison of class separability on benchmark datasets using the class-associated space of our method VS. the attribute latent space of ICAM-reg. The classification accuracy (mean$\pm$std) of ten-fold cross validation using a random forest classifier is reported.}
	\label{tsne-classification-analysis}
	\centering
		\scalebox{1}{
		\begin{tabular}{lccccc}
			\toprule
			Datasets    &OCT   &Brain Tumor1  &Brain Tumor2  &Chest X-rays  &Human Face \\
			\midrule
			ICAM-reg   &0.596$\pm$0.036   &0.843$\pm$0.040 &0.958$\pm$0.016 &0.886$\pm$0.030 &0.880$\pm$0.011  \\
			\midrule
			CAE (ours)  &\textbf{0.956$\pm$0.012}  &\textbf{0.970$\pm$0.015} &\textbf{0.974$\pm$0.010} &\textbf{0.909$\pm$0.025} &\textbf{0.964$\pm$0.005}  \\
			
			\bottomrule
		\end{tabular}
	}
\end{table*}

\begin{figure}
	\centering
	\includegraphics[scale=0.128]{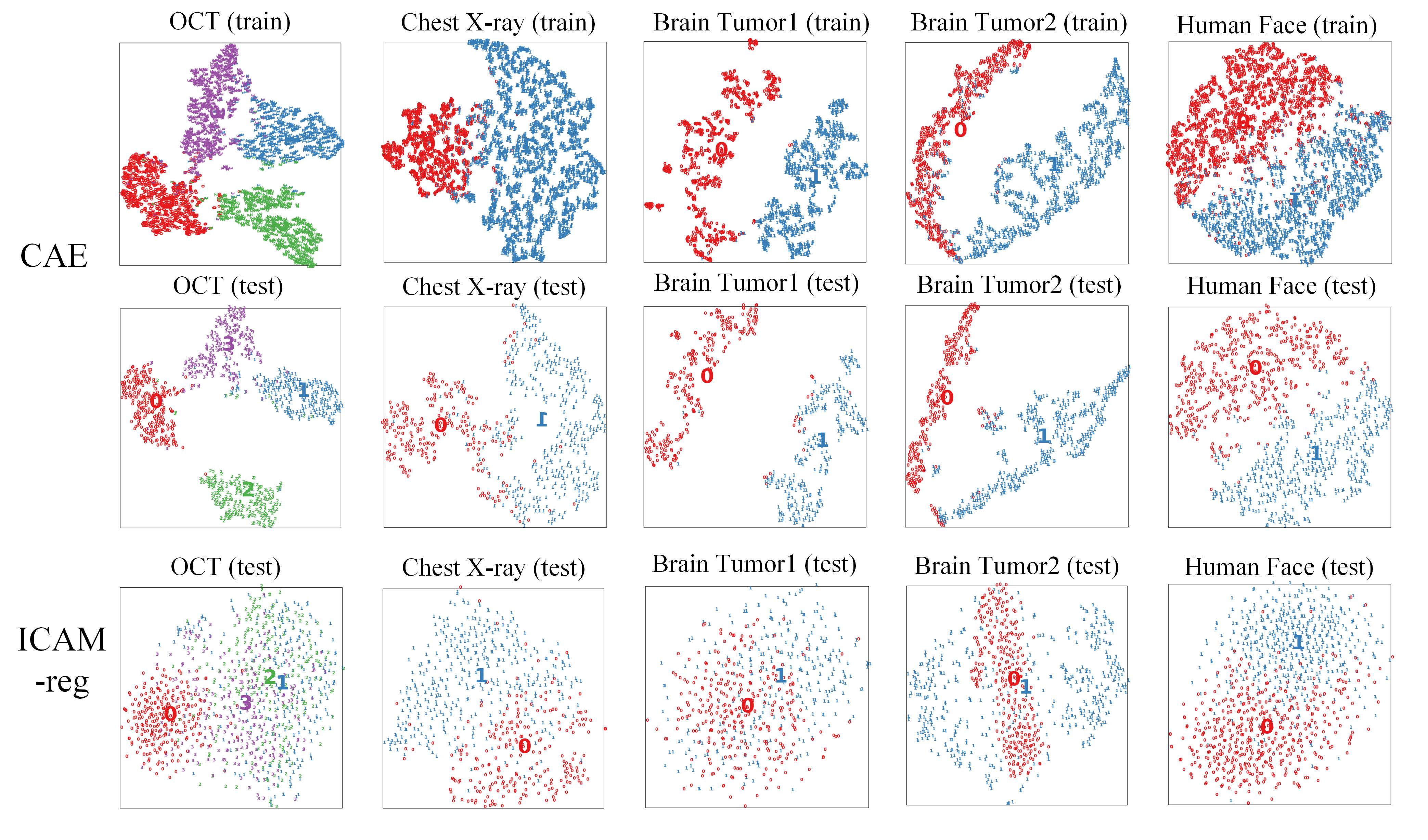}
	\caption{The t-SNE visualizations of manifold topology in the class-associated space extracted from the training (top row) and test (middle row) data using CAE versus the attribute latent space of ICAM-reg for the test data (bottom row) on benchmark datasets. Numbers with different colors represent different classes. For four medical datasets, red ‘0’ refers to normal cases, while colored numbers '1'-'3' refer to different abnormal sub-classes. For Human Face dataset, red ‘0’ refers to female cases while blue ‘1’ represents male cases.}
	\label{tsne-all-data}
\end{figure}

To visually demonstrate the effect of classification patterns for both methods, we adopted t-SNE to visualize our class-associated manifold and the attribution latent space of ICAM-reg on two-dimensional planes. The resulting figures were depicted in Fig. \ref{tsne-all-data}. We observed that samples from images with different classes were not only clearly separated apart, as expected, for both the training and test data, but also exhibited a similar distribution between training and test data, justifying that our method generalized to out-of-sample distributions, and the manifold was stable for global explanation (e.g., we can further correlate the features with external clinical variables, and/or merge multiple datasets, for knowledge discovery). In contrast, for ICAM-reg the test datasets were poorly separated. 
Furthermore, we see that our CAE method preserved the topological details of the datasets that showed analogues between training and test data. For example, in the OCT dataset, the three abnormal subclasses (CNV, DRUSEN, DME) showed similar spatial relations in training and test data. Furthermore, without supervised information, the fact that 
DRUSEN samples were positioned adjacent to the path from NORMAL to CNV, were agreed with existing medical knowledge that the development of DRUSEN may transit into CNV [78], [79]. Contrarily, the attribute latent space generated by ICAM-reg did not maintain any topological details but all classes were degraded into a Gaussian-like distribution.

The superiority of our method over previous counter-factual generation approaches is closely connected with the previously described learning techniques. Firstly, with BBCFE, we can obtain a significantly more optimal feature sets than the disentangling techniques used by previous algorithms (including ICAM-reg) which resulted in local optima due to their greedy-algorithm nature. Secondly, the introduction of both loss function equation \ref{con:loss2} and equation \ref{con:loss3} ensured the locally smoothness, and further the so-called homeomorphic (topology-maintaining) property of the embedding manifold. Although ICAM-reg also adopted a loss function analogue to equation \ref{con:loss2}, this restriction alone could not guarantee the locally smoothness of the mapping. Despite that equation \ref{con:loss3} did not operate on the class-associated latent space directly, its absence caused the non-smoothness of the individual feature space, which then in turn led to drifts in the peered class-associated code space and distortions of the data topology. Hence, our method not only achieved better class-separation in the latent space, but also successfully distilled the data topological structures that are conforming to domain knowledge, thereby affirming the capability of our model in discovering latent knowledge and data patterns adhering to global rules. 

\subsection{Semantically-Relevance Explanations achieved in the Global Class-Associated Manifold}
We proposed that the class-associated manifold provided a global map of decision pattern which can guide explanations on the entire dataset. To ensure this, the manifold must be semantic meaningful in the sense that each point in the manifold should be mapped to some feature patterns that are related to the classifier task and are common across samples. To validate this, we carried out experimental analyses to explore the nature of the obtained manifolds.
\subsubsection{Semantic Pervasiveness of the Class-Associated Manifold}
A successful extraction of class-associated feature means that when the extracted class-associated code is assigned to the other samples, the class assignments, judged by the target black-box classifier, should also be transferred to the modified samples. Moreover, this transferring should be pervasive, meaning that the class transmission should be applicable to the vast majority of samples. To validate this, we performed a class-swapping experiments on all datasets. For the samples in the test sets (whose codes were not used during model building), both the individual and class-associated codes were calculated and the class-associated codes from different classes were swapped to generate new samples. The proportions of the successful class re-assignments (judged by the trained classifier) on all datasets were shown in TABLE \ref{gen-acc}. We see that our method achieved high success rate of class re-assignments on all datasets. In contrasts, previous counter-factual methods such as ICAM-reg, performed poorly on the test dataset despite that they have enforced successful class transmission on the training samples. Meaning that the latent space learnt by ICAM-reg did not have semantic pervasiveness, while our method did.

\begin{table*}
	
	\caption{THE SUCCESS RATE CLASS RE-ASSIGNMENTS ON TEST DATA FROM BENCHMARK DATASETS USING SWAPPED CLASS-ASSOCIATED VERSUS ATTRIBUTE LATENT CODES.}
	\label{gen-acc}
	\centering
	\scalebox{1}{
		\begin{tabular}{lccccc}
			\toprule
			Datasets    &OCT   &Brain Tumor1  &Brain Tumor2  &Chest X-rays  &Human Face \\
			\midrule
			ICAM-reg  &60.0\%  &37.3\% &15.7\% &38.0\% &82.2\% \\
			\midrule
			CAE (ours)   &\textbf{92.3\%}  &\textbf{88.8\%} &\textbf{91.8\%} &\textbf{91.8\%} &\textbf{98.5\%} \\
			
			\bottomrule
		\end{tabular}
	}
\end{table*}

\begin{table*}[htbp]
	\caption{Comparison of the average time required to generate one saliency map using different XAI algorithms.}
	\label{time-cal}
	\centering
		\scalebox{1}{
			\begin{tabular}{lcccccccccc}
				\toprule
				Methods    &LIME  &Gradcam &Fullgrad &Simple Fullgrad &Smooth Fullgrad &ICAM-reg &LAGAN &TSCAM &StyLEx  &Ours \\
				\midrule
				Computation cost (ms)   &18900 &114 &105 &110 &894 &48 &28 &27 &222840 &\textbf{25}  \\
				
				\bottomrule
			\end{tabular}
		}
	\end{table*}

\subsubsection{Semantic Coherancy of the Class-Associated Codes}

\begin{figure}
	\centering
	\includegraphics[scale=0.18]{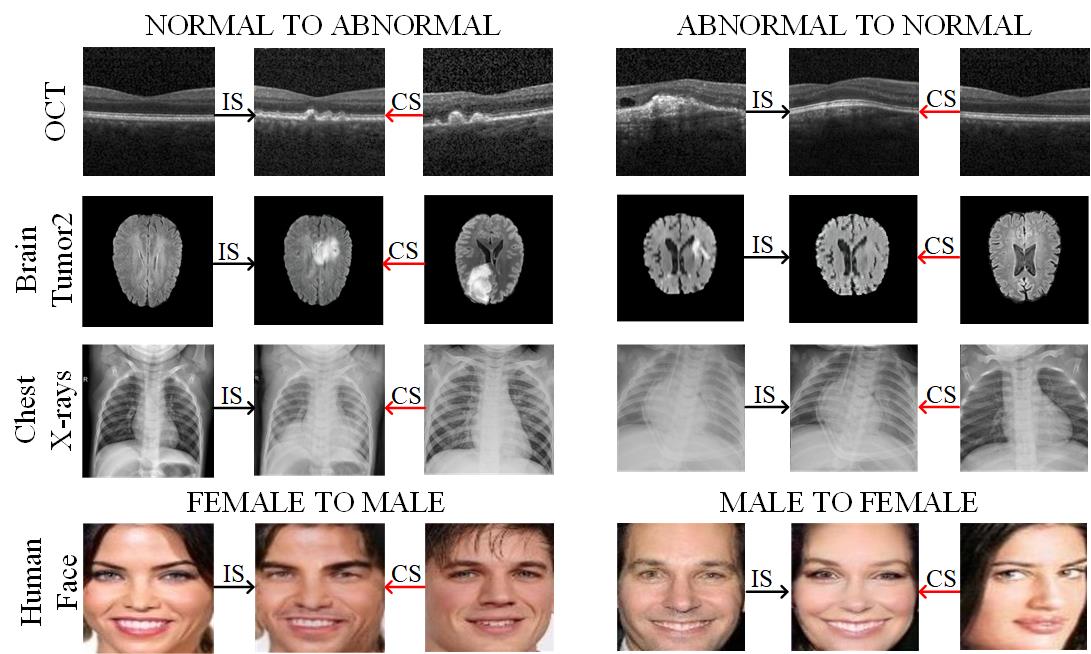}
	\caption{Synthetic cases (at the middle) based on the combination of the individual (IS) and class-associated (CS) codes using CAE on the OCT, Brain Tumor2, Chest X-rays and Human Face datasets.}
	\label{synthetic-cases}
\end{figure}

\begin{figure}
	\centering
	\includegraphics[scale=0.14]{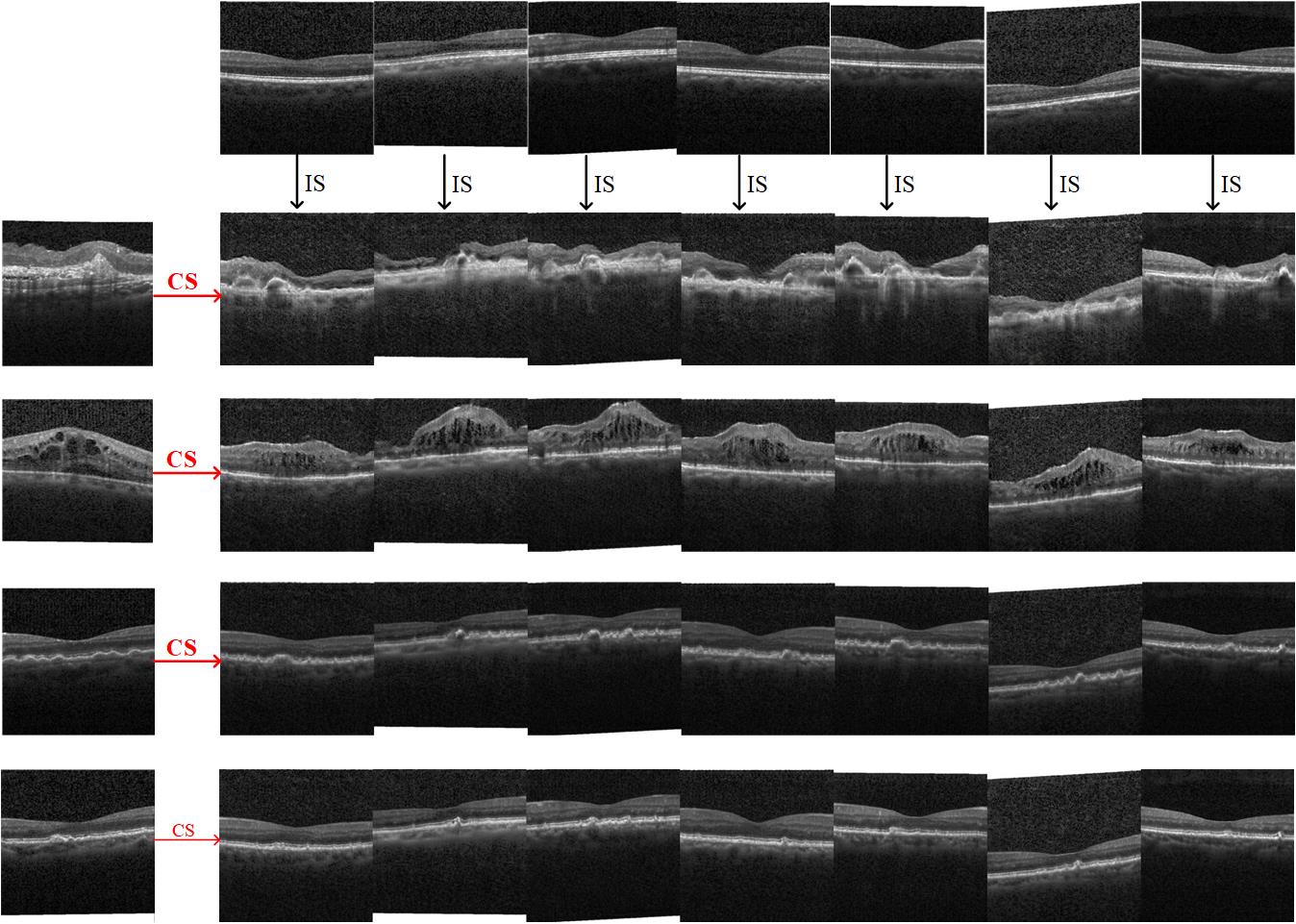}
	\caption{Cases that combined the class-associated codes (extracted from the images in the left column) with different individual codes (extracted from the images in the top row) for synthesizing new samples using CAE on the OCT dataset.}
	\label{semantic-all2}
\end{figure}

We verified that class-associated codes preserved visual characteristics associated with different classes in various types of images, including retinal images, brain images, chest X-rays and human face images. We provided several samples as depicted in Fig. \ref{synthetic-cases}. All generated samples (shown in the middle) retained the individual/identity features of images whose IS codes were used, including contours, shapes, sizes, structures (such as skeletal structures for brain images) and facial expressions (human face), while preserved class-associated features of the images whose CS codes were used, including wavy texture (OCT), white block-like objects (brain tumor), cloud-like, patchy high-density shadows (chest X-rays), and beards, eye shadow, eyebrow thickness and length, lip and skin redness (human face). We further demonstrated that the class-associated codes were connected to visual features that could be shared across different samples. As an example, we carried out experiment on the OCT test dataset by extracting the class-associated codes of one sample and combining it with 7 different individual backgrounds. 
The experiment was repeated on multiple samples and the visual results were demonstrated in Fig. \ref{semantic-all2}. We can see that similar class-associated features were well preserved when different background images were combined in these cases, further demonstrating the powerful semantic coherancy of the class-associated codes. 

\subsubsection{Smoothness of the Class-Associated Coding Manifold}
We proposed that our method achieves a locally smooth mapping on the encoding space. This is important because during training the codes were only extracted from sample data which are discrete and limited in number. It is important that the embedding can be generalized to the out-of-sample regions. To verify this, we adopted SMOTE (Synthetic Minority Over-Sampling Technique) to generate a set of 2000 resampled new codes (which were the convex combination of existing class-associated codes and distributed on the surface of the manifold contour) for each category in the OCT test data set, then combining the new class-associated codes with one individual code to generate new synthetic samples. We calculated the ratios of the new generated samples with successful class assignments, which were 93.44\% (Normal), 97.20\% (CNV), 94.11\% (DME), 97.58\% (DRUSEN) respectively. The wrongly assigned data may be caused by imprecise manifold contour caused by limited data. Furthermore, we show that when moving along a random linear path connecting two classes in the class-associated manifold, the predicted classification probabilities by the classifier vary in a continuous and monotonous manner, as is shown in Fig. \ref{oct-drag-gen}.(b). Therefore, we are confident that our class-associated domain is continuous and separable, which means most of the class-associated codes falling into the same subspace are with the same class association and the synthetic samples based them can be attributed to the same category. This further justifies the semantic relevance of the class-associated manifold.

\subsubsection{Global Knowledge Visualization by Exploring the Class-Associated Manifold}
\begin{figure}
	\centering
	\includegraphics[scale=0.127]{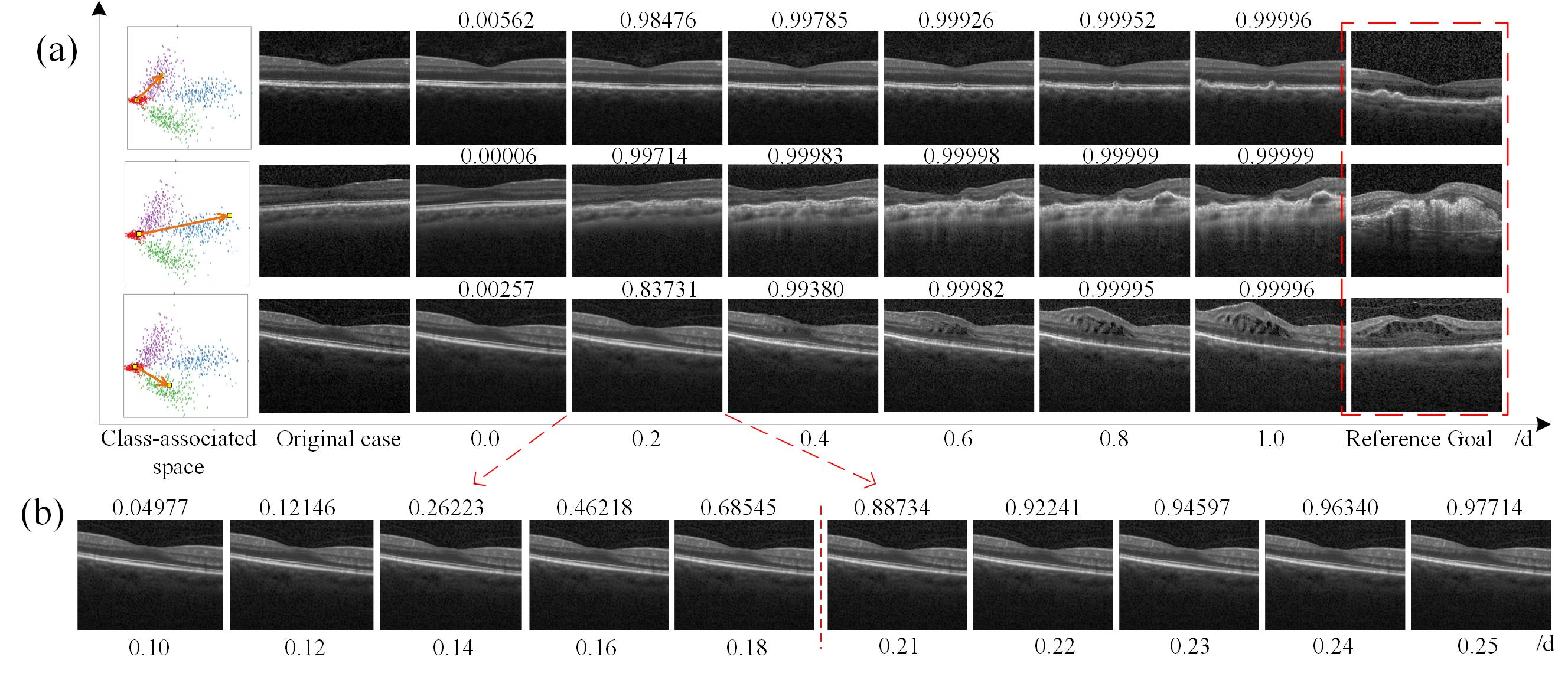}
	\caption{Generated images based on dragged class-associated codes (indicated by the brown rows) with different relative distances from the origin. The numbers on the images represent the probability that the classifier predicts the target class.}
	\label{oct-drag-gen}
\end{figure}

With the successful creation of the class-associated manifold, we gained a full picture of the class transition patterns across the entire dataset and the long-distance behavior of the black-box model, which has not been achieved in previous approached. In addition to creating more precise instance-based explanations, this method is also useful in visualizing the global decision pattern by creating multiple paths inside the manifold, and analyzing the visual features of images generated along different paths using the same individual sample as a template. We demonstrated this character of our methods on the OCT dataset as depicted in Fig. \ref{oct-drag-gen}.(a). We randomly selected two samples of different classes (one normal and the other abnormal) and created a linear path between them in the class-associated space. The individual code of the normal sample was used in combination with different class-associated codes sampled evenly along the path, resulting in a series of synthetic images. We observed that as the class-associated codes moved from the normal group toward the abnormal one, the pathological features of the corresponding synthetic images evolved with similar lesion characters, which vividly uncovered the underlying classification/pathological rules. By setting up various paths in the class-associated space, a set of different classification rules with clear medical meanings can be conveniently explored, enabling us to reveal and visually explain the global knowledge across entire dataset, encompassing relations between classes or samples, and global rules of class transferring. By observing the predictions generated by a black-box classifier for images generated along different paths, we can gain a more global understanding of the classifier’s long-range behavior. 

\subsection{Analysis of Computation Burden for Saliency Maps Generation} 
To objectively analyzing the computational burden of our algorithm, we conducted a set of comparative experiments. Using PyTorch software libarary V1.12.1 on a single NVIDIA TITAN Xp 12GB and Intel(R) Xeon(R) CPU E5-2650 v4@ 2.20GHz, we calculated the average time required for different algorithms to obtain the saliency map of one brain image (100 brain images were tested), and the experimental results were presented in TABLE \ref{time-cal}. It can be seen our method needs lowest time for generating one saliency map, demonstrating the high efficiency and low computation burden of our method. Besides, we calculated the training time and compared it with other methods that training networks is also required, as presented in TABLE \ref{training-time}. As a result, Our method has the least training time on all five datasets.

\begin{table}

	\caption{Comparison of the training time required for different algorithms.}
	\label{training-time}
	\centering
		\scalebox{1}{
			\begin{tabular}{lcccc}
				\toprule
				Methods    &ICAM-reg  &LAGAN   &StyLEx  &Ours \\
				\midrule
				OCT (h)    &27.3 &55.7 &46.3 &\textbf{17.5}   \\
				\midrule
				Brain Tumor1 (h)    &2.8 &3.8 &20.0 &\textbf{2.6} \\
				\midrule
				Brain Tumor2 (h)    &7.6 &14.0 &33.6 &\textbf{4.2}   \\
				\midrule
				Chest X-rays (h)   &8.2 &13.6 &31.2 &\textbf{4.6}   \\
				\midrule
				Human Face (h)  &20.3 &109.0 &49.5 &\textbf{18.7}   \\
				
				\bottomrule
			\end{tabular}
		}
\end{table}

\section{Conclusion and Future works}
With the increasing use of deep learning models in data management and analysis, the threat of undetected model biases or defects is surging. In image management and analysis tasks where a deep-learning image classifier is often involved, it would be important to validate whether the image classifier (can be a retrieval engine, an auto-annotation module, an image interpreter, a question answering agent, or a data mining model, etc.) learned the correct classification knowledge. For this purpose, an explainable AI that can precisely connect annotations/labels with image-level features while at the same time visualize the global decision behavior of the classifier is favored. In this paper, to effectively address the explainability challenge in black-box image classifiers, we have proposed a representation learning technique called Class Association Embedding, which efficiently embeds the class-associated features of the dataset into a low-dimensional manifold that is shown to coincide with domain knowledge and emulate the behavior of the classifier with high precision. We went on to develop an explainable learning framework based on CAE, where the classification rules can be visually explained by actively manipulating the class-associated code of any given sample to perform continuous and intentional modifications on the sample for changing its class-associated features but maintaining its individual characters. This technology achieved global decision knowledge distillation and sample-specific attribution at the same time, which combine the advantages of both global and local explanation approaches, and cannot be achieved by previous XAI methods. We carried out benchmark experiments showing that our method produced more detailed and accurate saliency maps for explaining image classifications compared with state-of-the-art methods, by skipping local optima traps with the aid of the global manifold learned from the dataset. Further analysis demonstrated that the class-associated manifold not only captured the semantically meaningful features of the images, but also learned the data pattern consistent with proven rules in reality on the datasets without supervised information, and also with desired properties of semantic consistency and class-distribution smoothness. For future study, we strive to develop methods for exploring the semantic power of the class-associated manifold to aid knowledge discovery in challenging AI tasks such as biomedical and scientific data analysis.

\section*{Acknowledgment}
This work was supported by the Strategic Priority Research Program of Chinese Academy of Sciences (grant no. XDB38050100), Shenzhen Science and Technology Program under grant no. KQTD20200820113106007, Shenzhen Key Laboratory of Intelligent Bioinformatics (ZDSYS20220422103800001), the National Natural Science Foundation of China under grant No. U22A2041.

\vspace{12pt}

\end{document}